\documentclass[sn-mathphys]{sn-jnl}% Math and Physical Sciences Reference Style
\usepackage{wrapfig}
\pdfoutput=1
\usepackage{graphicx}
\usepackage{hyperref}
\hypersetup{hypertex=true,colorlinks=true,linkcolor=blue,anchorcolor=blue,citecolor=blue}
\usepackage{doi}
\usepackage{amsmath}
\usepackage[numbers,sort&compress]{natbib}
\jyear{2022}%

\theoremstyle{thmstyleone}%
\theoremstyle{thmstyletwo}%

\theoremstyle{thmstylethree}%

\usepackage{caption}
\usepackage{multirow}
\usepackage{makecell}
\usepackage{amsmath}
\usepackage{tabularx}

\begin{document}

\title[Article Title]{SAI: Solving AI Tasks with Systematic Artificial Intelligence in Communication Network}


\author[1]{\fnm{Lei Yao}}

\author*[1,2]{\fnm{Yong Zhang}}\email{yongzhang@bupt.edu.cn}

\author[1]{\fnm{Zilong Yan}}

\author[1]{\fnm{Jialu Tian}}

\affil[1]{\orgdiv{School of Electronic Engineering}, \orgname{Beijing University of Posts and Telecommunications}, \orgaddress{\city{Beijing}, \postcode{100876}, \state{China}}}

\affil[2]{\orgdiv{Beijing Key Laboratory of Work Safety Intelligent Monitoring}, \orgname{Beijing University of Posts and Telecommunications}, \orgaddress{ \city{Beijing}, \postcode{100876}, \state{China}}}


\abstract{
In the rapid development of artificial intelligence, solving complex AI tasks is a crucial technology in intelligent mobile networks. Despite the good performance of specialized AI models in intelligent mobile networks, they are unable to handle complicated AI tasks. To address this challenge, we propose Systematic Artificial Intelligence (SAI), which is a framework designed to solve AI tasks by leveraging Large Language Models (LLMs) and JSON-format intent-based input to connect self-designed model library and database. Specifically, we first design a multi-input component, which simultaneously integrates Large Language Models (LLMs) and JSON-format intent-based inputs to fulfill the diverse intent requirements of different users. In addition, we introduce a model library module based on model cards which employ model cards to pairwise match between different modules for model composition. Model cards contain the corresponding model's name and the required performance metrics. Then when receiving user network requirements, we execute each subtask for multiple selected model combinations and provide output based on the execution results and LLM feedback. By leveraging the language capabilities of LLMs and the abundant AI models in the model library, SAI can complete numerous complex AI tasks in the communication network, achieving impressive results in network optimization, resource allocation, and other challenging tasks. 
}




\maketitle

\section{Introduction}\label{sec1}

%The Introduction section, of referenced text \cite{bib1} expands on the background of the work (some overlap with the Abstract is acceptable). The introduction should not include subheadings.

%Springer Nature does not impose a strict layout as standard however authors are advised to check the individual requirements for the journal they are planning to submit to as there may be journal-level preferences. When preparing your text please also be aware that some stylistic choices are not supported in full text XML (publication version), including coloured font. These will not be replicated in the typeset article if it is accepted. 
In recent years, with the development of 5G technology and computing power network, significant developments have occurred in network computing power and communication networks. This evolution has profoundly altered the way networks are constructed and AI tasks are executed. Additionally, with the significant breakthroughs achieved by LLM in human-like intelligence, handling complex AI tasks based on intent has become feasible. Recently, the key method of multi-agent collaboration based on LLM \cite{shen2023hugginggpt,hong2023metagpt,tang2023towards}, such as HuggingGPT \cite{shen2023hugginggpt}, MetaGPT \cite{hong2023metagpt}, and so on, is to use LLM as a controller to handle complex task translation and planning. Then different agents are employed to execute various modules to effectively complete all kinds of complex tasks. These methods have good Performance on complex tasks. However, their performance is limited by the LLM model. Moreover, to our knowledge, there is no literature that has applied this approach to complex AI tasks in mobile networks.

In mobile networks, Intent-Based Networking (IBN) \cite{leivadeas2022survey} is a network paradigm that automates network configuration based on intent analysis. With the rise of IBN, Software-Defined Networking (SDN) \cite{benzekki2016software} has achieved significant success. Current approaches based on intent networks \cite{kiran2018enabling,mahtout2020using,dzeparoska2021towards} primarily focus on intent classification and multi-intent conflicts. Intent classification typically categorizes intents into two groups: end users and network operators. Then they provide different intent inputs for different users. Multi-intent conflicts focus on how to execute different intents simultaneously during the intent input process. However, These methods are generally applicable only in simple environments and cannot handle complex AI tasks in mobile networks.

To address these challenges, we propose SAI framework. The framework is designed to solve AI tasks by leveraging Large Language Models (LLMs), JSON-format intent-based input and self-designed model library and database. Specifically, we simultaneously integrates Large Language Models (LLMs) and JSON-format intent-based inputs to fullfill the diverse intent requirements of different users. In addition, we introduce a model library module based on model cards which employ model cards to pairwise match between different modules for model composition. Then when receiving user network requirements, we execute each subtask for multiple selected model combinations and provide output based on the execution results and LLM feedback. The main contributions of this paper are summarized as follows:

\begin{itemize}
%\begin{itemize}
\item[$\bullet$] To solve complex AI tasks in communication networks, We propose SAI, which uses multi-input component to interact with self-designed model libraries, databases. To the best of our knowledge, this is the first systematic artificial intelligence approach to solve complex tasks in communication networks.
\end{itemize}
\begin{itemize}
\item[$\bullet$] We propose the multi-input component, which combines LLM input with intention input based on JSON format to fulfill the requirements of different people. In addition, we design model cards-based model library. Specifically, model cards contain the corresponding model's name and the network performance metrics. Then we employ model cards to pairwise match between different modules for model composition.
\end{itemize}

The remainder of this paper is organized as follows. A brief review of related work is provided in Section \ref{sec2}. In Section \ref{sec3}, we detail the Framework for Systematic Artificial Intelligence. Section \ref{sec4} comes to concluding remarks.

%\section{Results}\label{sec2}

%Sample body text. Sample body text. Sample body text. Sample body text. Sample body text. Sample body text. Sample body text. Sample body text.
\section{Related Work}\label{sec2}
\subsection{Autonomous Agents}\label{subsec1}
Autonomous agents are commonly regarded as a key technology in the artificial general intelligence (AGI). Their autonomous governance and decision-making have attracted significant interest. Previous approaches generally relied on reinforcement learning to interact with the environment for achieving autonomous governance. Volodymyr Mnih et al.\cite{mnih2015human} demonstrate that the deep Q-network agent, receiving only the pixels and the game score as inputs. And this network bridges the divide between high-dimensional sensory inputs and actions, resulting in the first artificial agent that is capable of learning to excel at a diverse array of challenging tasks. John Schulman et al. \cite{schulman2017proximal} propose a new family of policy gradient methods for reinforcement learning, which alternate between sampling data through interaction with the environment, and optimizing an "agent" objective function using stochastic gradient ascent. Tuomas Haarnoja et al. \cite{haarnoja2018soft} propose soft actor-critic, an off-policy actor-critic deep RL algorithm based on the maximum entropy reinforcement learning framework. In this framework, the actor aims to maximize expected reward while also maximizing entropy. That is, to succeed at the task while acting as randomly as possible. These reinforcement learning-based methods assume that the agent is a simple policy function and are limited to specific environments. These assumptions differ from human decision-making processes, and previous approaches were unable to interact with humans or applied in open-domain environments.  

With the tremendous success of large language models (LLMs), it has shown immense potential in attaining human-like intelligence. To address the challenges of autonomous agents' limited applicability in complex environments, Yongliang Shen et al. \cite{shen2023hugginggpt} propose HuggingGPT, a framework that leverages LLMs (e.g.,ChatGPT) to connect various AI models in machine learning communities (e.g.,Hugging Face) to solve AI tasks. And Sirui Hong et al. \cite{hong2023metagpt} introduce MetaGPT, an innovative framework that incorporates efficient human workflows as a meta programming approach into LLM-based multi-agent collaboration. Shujian Zhang et al.  \cite{zhang2023automl} present the AutoML-GPT, which employs GPT as the bridge to diverse AI models and dynamically trains models with optimized hyperparameters. AutoML-GPT dynamically takes user requests from the model and data cards and composes the corresponding prompt paragraph. Multi-agent collaboration based on LLM can complete numerous complex AI tasks. However, it cannot avoid the LLM of hallucination problem \cite{rawte2023survey}. This could be critical particularly in mobile network scenarios.
\subsection{Intent-Based Network(IBN)}\label{subsec2}
Intent-Based Networking (IBN) is a network paradigm to Automatically manage network configuration based on intent \cite{leivadeas2022survey}. Recently, intent-based methods with chatbot interfaces have emerged to simplify intent translation and incorporate user feedback. Tools like iNDIRA \cite{kiran2018enabling} utilize NLP to create semantic RDF graphs, which are translated into network commands. EVIAN \cite{mahtout2020using} extends iNDIRA by using RASA for the chatbot interface and a hierarchy of RDFs for intent translation. LUMI \cite{jacobs2021hey} employs Google Dialogflow and learning methods to translate user intents into Nile intents, which are compiled into programs for network configuration changes. \cite{dzeparoska2021towards} define a formal policy framework which allows modeling policies at different levels of abstraction, including utility, goal, and event-condition-action (ECA) policy types, and enables conflict detection and resolution across abstraction layers. At a minimum, a policy should contain a set of resources to which an action is to be applied, while considering a set of associated constraints. These intent networks using natural language as input satisfies end users' network demands. However, the hallucination problem associated with LLM (Language Model Learning) could potentially be critical in mobile network . In addition, the current approach based on intent networks don’t consider selecting appropriate models for complex AI tasks.

\section{Methodology}\label{sec3}

The SAI is a multi-agent collaborative system designed to complete complex AI tasks in mobile networks. It is composed of multiple inputs, including LLM inputs and JSON-formatted intent inputs, a self-designed model library and database. Its workflow consists of five stages: multiple inputs component, task translation and planning, model selection strategy, task execution, final output and response feedback, as shown in Figure 1. Given a end user requirements or a network operator specialized intent input, SAI automatically accomplish a variety of complex AI tasks using different models. In the subsequent sections, we will delve into the design of each stage.

\begin{figure}[h]%
\centering
\includegraphics[width=\textwidth]{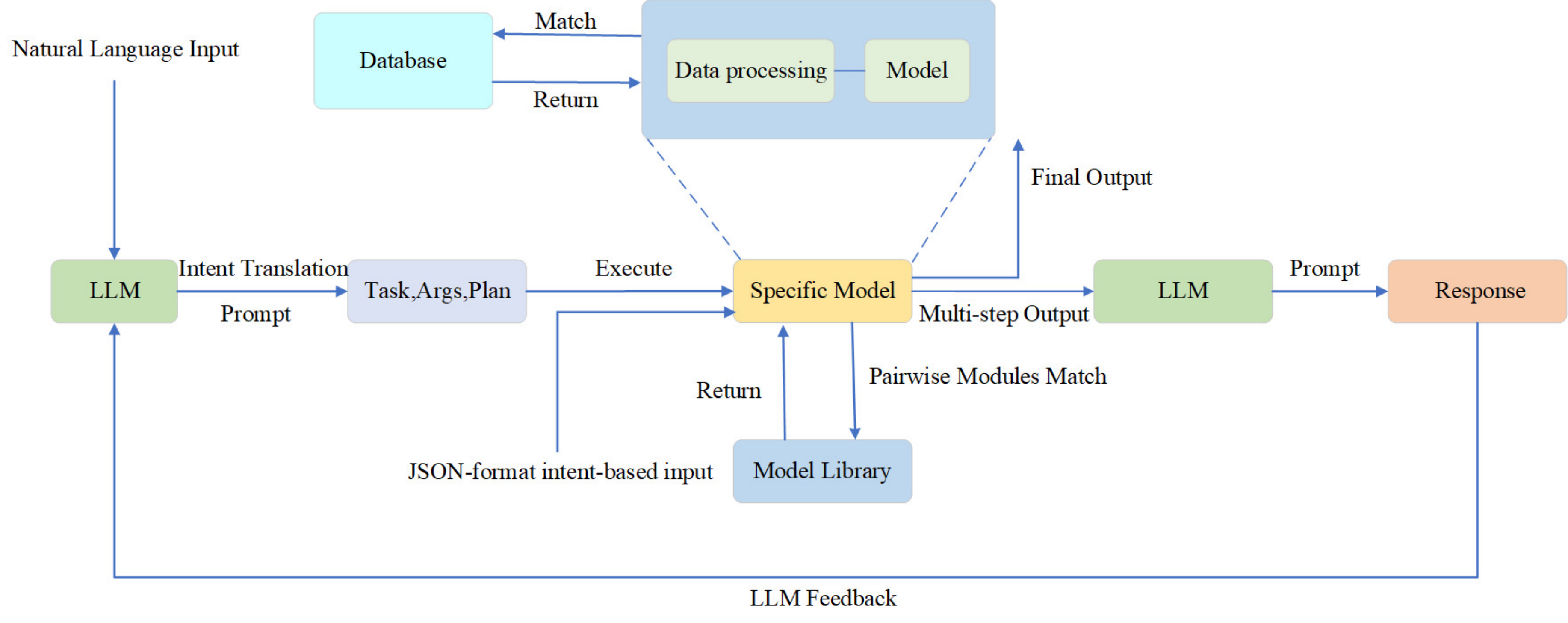}
\caption{The Framework for Systematic Artificial Intelligence}\label{fig1}
\end{figure}

We propose a new neural network structure named DLA-GCN. As shown in Fig. \ref{fig1}, the model consists of the DLGCN, DATL , and NMPL components. These approaches learn the spatial structural properties by combining the DLGCN and NMPL and employs the DATL algorithm to learn and transfer similar features of dynamic graphs at adjacent time steps. Note that nodes in this section dynamically evolve. In other words, the number of nodes in each time step is different. Therefore, after feature extraction at $t$, we add 0 to match the dimension in adjacent time steps $t$ and $t+1$. This approach avoids dimension mismatches and prevents the occurrence of the multi-zero matrix.  

\subsection{Multiple Inputs Component}\label{subsec1}
The multi-input component primarily consists of two parts: LLM natural language input and intent input based on the JSON format. End users tend to interact with models using natural language to complete complex tasks and thereby the use of LLM is necessary. However, due to the inherent ambiguities in natural language and hallucination problem in LLM, we consider incorporating integrated feedback of LLM to enhance the accuracy of its translation and planning, which will be discussed in Section 3.5. Meanwhile, for network operators, we provide unambiguous intent input based on the JSON format, directly conveying the true intent to the network to ensure the correct execution of network requirements.

\subsection{Task Translation and Planning}\label{subsec2}
To solve the complex AI tasks in mobile networks, it is necessary to coordinate multiple sub-tasks. After processing the LLM input, we employ LLM to analyze user intentions and decompose them into a set of structured tasks. Furthermore, we also require the LLM to determe dependencies and execution orders for these decomposed tasks to establish their connections by the re-prompting method \cite{raman2022planning}. Currently, supported task relationships include single tasks, chained tasks, and tree-structured tasks. To ensure effective and accurate task planning by LLM, Our models employs a feedback-driven prompt-based design, including specification-based instructions, demonstration-based parsing, and parsing optimization with task output feedback. The feedback-driven parsing optimization will be elaborated upon in Section 3.5. 

In addition, to support the requirements of multi-turn conversations, we incorporate chat logs into prompts through additional instructions. Specifically, to assist in task planning, chat history can be used as {{chat logs}} to track resources mentioned by the user and incorporate them into subsequent task planning. This design enables our model to better manage context and more accurately solve AI tasks in multi-turn dialogues.

\subsection{Model Selection Strategy}\label{subsec3}

Due to the complexity of the AI tasks, after completing task planning, we need to select model combinations from various candidate models that can satisfy the intent requirements. To address this, we first introduce a model card-based model library. Specifically, in the model library, we have different candidate models and each model is assigned a corresponding model card. Under the assumption of a stable network environment, each model card includes the model's name and various network performance metrics, such as latency and resource utilization.

Then during the model selection process, we utilize a model matching mechanism to generate multiple suitable model combinations based on intent requirements. Specifically, we perform pairwise matchs among the model cards from different model libraries based on the performance metrics specified in the intent requirements to fulfill the overall demand. The method of model cards not only reduces the complexity of model matching but also provide the foundation for potential network dynamics in the future.

\subsection{Task Execution}\label{subsec4}

Once a specific model is assigned to a parsed task, the next step is to execute the task, which involves model inference. At this stage, we propose a data card-based database and a unified input-output section for model libraries to facilitate model input. Specifically, the database contains all the required data and their corresponding data cards. Data cards include data names and key attributes, and the data can be used simply by employing the data card.

After selecting a specific model, the model directly uses the database through the model library's unified input-output component. The data is input to the model in a standardized format, and the output is also standardized. Additionally, to ensure the sequential execution of tasks, we consider using re-prompting \cite{raman2022planning} to dynamically specify task dependencies and resources.

Furthermore, for sub-tasks within a tree-like structure, we execute these tasks in parallel when they have no resource dependencies between them, thus further enhancing inference efficiency.

\subsection{Final Output and LLM Feedback}\label{subsec5}

After completing all tasks, we need to generate the final result and LLM feedback. For the final output, we simply show the results of the model combinations that fulfill the requirements and their corresponding execution performance . To optimize LLM translation and task planning in detail, we integrates all information from the first three stages (task planning, model selection, and task execution) into a formatting summary for this stage. This summary includes the planned task list, the models selected for task execution, and the inference results from these models. These integrated outputs are presented in a structured prompting format. 

Then scores is assigned to the results of these three stages, particularly focusing on explaining the reasons for any errors. Finally, this feedback is provided to LLM to enhance its performance.

\section{Conlusion}\label{sec4}
To solve complex AI tasks in communication network, We propose Systematic Artificial Intelligence (SAI) by leveraging Large Language Models (LLMs) and JSON-format intent-based input to connect self-designed model library and database. Specifically, we first design a multi-input component, which simultaneously  integrates Large Language Models (LLMs) and JSON-format intent-based inputs to fullfill the diverse intent requirements of different users. In addition, we introduce a model library module based on model cards which employ model cards to pairwise match between different modules for model composition. Model cards contain the corresponding model's name and the required performance metrics. Then when receiving user network requirements, we execute each subtask for multiple selected model combinations and provide output based on the execution results and LLM feedback. By leveraging the language capabilities of LLMs and the abundant AI models in the model library, SAI can complete numerous complex AI tasks in the communication network, achieving impressive results in network optimization, resource allocation, and other challenging tasks.





%\bibliographystyle{sn-basic}
%\bibliographystyle{spbasic}
\bibliographystyle{sn-basic}
\bibliography{a} 


\begin{thebibliography}{15}
\ifx \bisbn   \undefined \def \bisbn  #1{ISBN #1}\fi
\ifx \binits  \undefined \def \binits#1{#1}\fi
\ifx \bauthor  \undefined \def \bauthor#1{#1}\fi
\ifx \batitle  \undefined \def \batitle#1{#1}\fi
\ifx \bjtitle  \undefined \def \bjtitle#1{#1}\fi
\ifx \bvolume  \undefined \def \bvolume#1{\textbf{#1}}\fi
\ifx \byear  \undefined \def \byear#1{#1}\fi
\ifx \bissue  \undefined \def \bissue#1{#1}\fi
\ifx \bfpage  \undefined \def \bfpage#1{#1}\fi
\ifx \blpage  \undefined \def \blpage #1{#1}\fi
\ifx \burl  \undefined \def \burl#1{\textsf{#1}}\fi
\ifx \doiurl  \undefined \def \doiurl#1{\url{https://doi.org/#1}}\fi
\ifx \betal  \undefined \def \betal{\textit{et al.}}\fi
\ifx \binstitute  \undefined \def \binstitute#1{#1}\fi
\ifx \binstitutionaled  \undefined \def \binstitutionaled#1{#1}\fi
\ifx \bctitle  \undefined \def \bctitle#1{#1}\fi
\ifx \beditor  \undefined \def \beditor#1{#1}\fi
\ifx \bpublisher  \undefined \def \bpublisher#1{#1}\fi
\ifx \bbtitle  \undefined \def \bbtitle#1{#1}\fi
\ifx \bedition  \undefined \def \bedition#1{#1}\fi
\ifx \bseriesno  \undefined \def \bseriesno#1{#1}\fi
\ifx \blocation  \undefined \def \blocation#1{#1}\fi
\ifx \bsertitle  \undefined \def \bsertitle#1{#1}\fi
\ifx \bsnm \undefined \def \bsnm#1{#1}\fi
\ifx \bsuffix \undefined \def \bsuffix#1{#1}\fi
\ifx \bparticle \undefined \def \bparticle#1{#1}\fi
\ifx \barticle \undefined \def \barticle#1{#1}\fi
\bibcommenthead
\ifx \bconfdate \undefined \def \bconfdate #1{#1}\fi
\ifx \botherref \undefined \def \botherref #1{#1}\fi
\ifx \url \undefined \def \url#1{\textsf{#1}}\fi
\ifx \bchapter \undefined \def \bchapter#1{#1}\fi
\ifx \bbook \undefined \def \bbook#1{#1}\fi
\ifx \bcomment \undefined \def \bcomment#1{#1}\fi
\ifx \oauthor \undefined \def \oauthor#1{#1}\fi
\ifx \citeauthoryear \undefined \def \citeauthoryear#1{#1}\fi
\ifx \endbibitem  \undefined \def \endbibitem {}\fi
\ifx \bconflocation  \undefined \def \bconflocation#1{#1}\fi
\ifx \arxivurl  \undefined \def \arxivurl#1{\textsf{#1}}\fi
\csname PreBibitemsHook\endcsname

%%% 1
\bibitem{shen2023hugginggpt}
\begin{botherref}
\oauthor{\bsnm{Shen}, \binits{Y.}},
\oauthor{\bsnm{Song}, \binits{K.}},
\oauthor{\bsnm{Tan}, \binits{X.}},
\oauthor{\bsnm{Li}, \binits{D.}},
\oauthor{\bsnm{Lu}, \binits{W.}},
\oauthor{\bsnm{Zhuang}, \binits{Y.}}:
Hugginggpt: Solving ai tasks with chatgpt and its friends in huggingface.
arXiv preprint arXiv:2303.17580
(2023)
\end{botherref}
\endbibitem

%%% 2
\bibitem{hong2023metagpt}
\begin{botherref}
\oauthor{\bsnm{Hong}, \binits{S.}},
\oauthor{\bsnm{Zheng}, \binits{X.}},
\oauthor{\bsnm{Chen}, \binits{J.}},
\oauthor{\bsnm{Cheng}, \binits{Y.}},
\oauthor{\bsnm{Zhang}, \binits{C.}},
\oauthor{\bsnm{Wang}, \binits{Z.}},
\oauthor{\bsnm{Yau}, \binits{S.K.S.}},
\oauthor{\bsnm{Lin}, \binits{Z.}},
\oauthor{\bsnm{Zhou}, \binits{L.}},
\oauthor{\bsnm{Ran}, \binits{C.}}, et al.:
Metagpt: Meta programming for multi-agent collaborative framework.
arXiv preprint arXiv:2308.00352
(2023)
\end{botherref}
\endbibitem

%%% 3
\bibitem{tang2023towards}
\begin{botherref}
\oauthor{\bsnm{Tang}, \binits{Z.}},
\oauthor{\bsnm{Wang}, \binits{R.}},
\oauthor{\bsnm{Chen}, \binits{W.}},
\oauthor{\bsnm{Wang}, \binits{K.}},
\oauthor{\bsnm{Liu}, \binits{Y.}},
\oauthor{\bsnm{Chen}, \binits{T.}},
\oauthor{\bsnm{Lin}, \binits{L.}}:
Towards causalgpt: A multi-agent approach for faithful knowledge reasoning via
  promoting causal consistency in llms.
arXiv preprint arXiv:2308.11914
(2023)
\end{botherref}
\endbibitem

%%% 4
\bibitem{leivadeas2022survey}
\begin{botherref}
\oauthor{\bsnm{Leivadeas}, \binits{A.}},
\oauthor{\bsnm{Falkner}, \binits{M.}}:
A survey on intent based networking.
IEEE Communications Surveys \& Tutorials
(2022)
\end{botherref}
\endbibitem

%%% 5
\bibitem{benzekki2016software}
\begin{barticle}
\bauthor{\bsnm{Benzekki}, \binits{K.}},
\bauthor{\bsnm{El~Fergougui}, \binits{A.}},
\bauthor{\bsnm{Elbelrhiti~Elalaoui}, \binits{A.}}:
\batitle{Software-defined networking (sdn): a survey}.
\bjtitle{Security and communication networks}
\bvolume{9}(\bissue{18}),
\bfpage{5803}--\blpage{5833}
(\byear{2016})
\end{barticle}
\endbibitem

%%% 6
\bibitem{kiran2018enabling}
\begin{barticle}
\bauthor{\bsnm{Kiran}, \binits{M.}},
\bauthor{\bsnm{Pouyoul}, \binits{E.}},
\bauthor{\bsnm{Mercian}, \binits{A.}},
\bauthor{\bsnm{Tierney}, \binits{B.}},
\bauthor{\bsnm{Guok}, \binits{C.}},
\bauthor{\bsnm{Monga}, \binits{I.}}:
\batitle{Enabling intent to configure scientific networks for high performance
  demands}.
\bjtitle{Future Generation Computer Systems}
\bvolume{79},
\bfpage{205}--\blpage{214}
(\byear{2018})
\end{barticle}
\endbibitem

%%% 7
\bibitem{mahtout2020using}
\begin{bchapter}
\bauthor{\bsnm{Mahtout}, \binits{H.}},
\bauthor{\bsnm{Kiran}, \binits{M.}},
\bauthor{\bsnm{Mercian}, \binits{A.}},
\bauthor{\bsnm{Mohammed}, \binits{B.}}:
\bctitle{Using machine learning for intent-based provisioning in high-speed
  science networks}.
In: \bbtitle{Proceedings of the 3rd International Workshop on Systems and
  Network Telemetry and Analytics},
pp. \bfpage{27}--\blpage{30}
(\byear{2020})
\end{bchapter}
\endbibitem

%%% 8
\bibitem{dzeparoska2021towards}
\begin{barticle}
\bauthor{\bsnm{Dzeparoska}, \binits{K.}},
\bauthor{\bsnm{Beigi-Mohammadi}, \binits{N.}},
\bauthor{\bsnm{Tizghadam}, \binits{A.}},
\bauthor{\bsnm{Leon-Garcia}, \binits{A.}}:
\batitle{Towards a self-driving management system for the automated realization
  of intents}.
\bjtitle{IEEE Access}
\bvolume{9},
\bfpage{159882}--\blpage{159907}
(\byear{2021})
\end{barticle}
\endbibitem

%%% 9
\bibitem{mnih2015human}
\begin{barticle}
\bauthor{\bsnm{Mnih}, \binits{V.}},
\bauthor{\bsnm{Kavukcuoglu}, \binits{K.}},
\bauthor{\bsnm{Silver}, \binits{D.}},
\bauthor{\bsnm{Rusu}, \binits{A.A.}},
\bauthor{\bsnm{Veness}, \binits{J.}},
\bauthor{\bsnm{Bellemare}, \binits{M.G.}},
\bauthor{\bsnm{Graves}, \binits{A.}},
\bauthor{\bsnm{Riedmiller}, \binits{M.}},
\bauthor{\bsnm{Fidjeland}, \binits{A.K.}},
\bauthor{\bsnm{Ostrovski}, \binits{G.}}, \betal:
\batitle{Human-level control through deep reinforcement learning}.
\bjtitle{nature}
\bvolume{518}(\bissue{7540}),
\bfpage{529}--\blpage{533}
(\byear{2015})
\end{barticle}
\endbibitem

%%% 10
\bibitem{schulman2017proximal}
\begin{botherref}
\oauthor{\bsnm{Schulman}, \binits{J.}},
\oauthor{\bsnm{Wolski}, \binits{F.}},
\oauthor{\bsnm{Dhariwal}, \binits{P.}},
\oauthor{\bsnm{Radford}, \binits{A.}},
\oauthor{\bsnm{Klimov}, \binits{O.}}:
Proximal policy optimization algorithms.
arXiv preprint arXiv:1707.06347
(2017)
\end{botherref}
\endbibitem

%%% 11
\bibitem{haarnoja2018soft}
\begin{bchapter}
\bauthor{\bsnm{Haarnoja}, \binits{T.}},
\bauthor{\bsnm{Zhou}, \binits{A.}},
\bauthor{\bsnm{Abbeel}, \binits{P.}},
\bauthor{\bsnm{Levine}, \binits{S.}}:
\bctitle{Soft actor-critic: Off-policy maximum entropy deep reinforcement
  learning with a stochastic actor}.
In: \bbtitle{International Conference on Machine Learning},
pp. \bfpage{1861}--\blpage{1870}
(\byear{2018}).
\bcomment{PMLR}
\end{bchapter}
\endbibitem

%%% 12
\bibitem{zhang2023automl}
\begin{botherref}
\oauthor{\bsnm{Zhang}, \binits{S.}},
\oauthor{\bsnm{Gong}, \binits{C.}},
\oauthor{\bsnm{Wu}, \binits{L.}},
\oauthor{\bsnm{Liu}, \binits{X.}},
\oauthor{\bsnm{Zhou}, \binits{M.}}:
Automl-gpt: Automatic machine learning with gpt.
arXiv preprint arXiv:2305.02499
(2023)
\end{botherref}
\endbibitem

%%% 13
\bibitem{rawte2023survey}
\begin{botherref}
\oauthor{\bsnm{Rawte}, \binits{V.}},
\oauthor{\bsnm{Sheth}, \binits{A.}},
\oauthor{\bsnm{Das}, \binits{A.}}:
A survey of hallucination in large foundation models.
arXiv preprint arXiv:2309.05922
(2023)
\end{botherref}
\endbibitem

%%% 14
\bibitem{jacobs2021hey}
\begin{bchapter}
\bauthor{\bsnm{Jacobs}, \binits{A.S.}},
\bauthor{\bsnm{Pfitscher}, \binits{R.J.}},
\bauthor{\bsnm{Ribeiro}, \binits{R.H.}},
\bauthor{\bsnm{Ferreira}, \binits{R.A.}},
\bauthor{\bsnm{Granville}, \binits{L.Z.}},
\bauthor{\bsnm{Willinger}, \binits{W.}},
\bauthor{\bsnm{Rao}, \binits{S.G.}}:
\bctitle{Hey, lumi! using natural language for $\{$intent-based$\}$ network
  management}.
In: \bbtitle{2021 USENIX Annual Technical Conference (USENIX ATC 21)},
pp. \bfpage{625}--\blpage{639}
(\byear{2021})
\end{bchapter}
\endbibitem

%%% 15
\bibitem{raman2022planning}
\begin{botherref}
\oauthor{\bsnm{Raman}, \binits{S.S.}},
\oauthor{\bsnm{Cohen}, \binits{V.}},
\oauthor{\bsnm{Rosen}, \binits{E.}},
\oauthor{\bsnm{Idrees}, \binits{I.}},
\oauthor{\bsnm{Paulius}, \binits{D.}},
\oauthor{\bsnm{Tellex}, \binits{S.}}:
Planning with large language models via corrective re-prompting.
arXiv preprint arXiv:2211.09935
(2022)
\end{botherref}
\endbibitem

\end{thebibliography}

\end{document}